\documentclass[a4paper,twocolumn]{article}
\usepackage[utf8]{inputenc}
\usepackage[english]{babel}
\usepackage[left=2cm, right=2cm, top=2.5cm, bottom=2.5cm]{geometry}
\usepackage{cite}

\usepackage{graphicx}
\graphicspath{ {./img/} }
\usepackage[dvipsnames]{xcolor}
\usepackage{tabularx}
\usepackage{float} 
\usepackage{multirow}
\usepackage{multicol}
\usepackage{array}
\usepackage{todonotes}
\usepackage{amssymb}
\usepackage{amsmath}

\usepackage{hyperref}
\hypersetup{
     colorlinks = true,
     linkcolor = green!60!black,
     anchorcolor = green!50!black,
     citecolor = green!60!black,
     filecolor = green!60!black,
     urlcolor = green!60!black,
     }
\usepackage[nameinlink,noabbrev]{cleveref}

\usepackage{tikz}

\usetikzlibrary{arrows}
\usetikzlibrary{positioning, shapes.geometric, shapes.misc}

\newcommand{\IoU}{\mathit{IoU}}

\newcommand{\mAP}{\mathit{mAP}}

\begin{document}

\newcolumntype{C}[1]{>{\centering\arraybackslash}p{#1}} 
\title{\textbf{YOdar: Uncertainty-based Sensor Fusion for Vehicle Detection with Camera and Radar Sensors}}
\date{}


\author
{Kamil Kowol,$^{1}$ Matthias Rottmann,$^{1}$ Stefan Bracke,$^{2}$ Hanno Gottschalk$^{1}$\\
\\
\normalsize{$^{1}$University of Wuppertal, School of Mathematics and Natural Sciences / IZMD / IMACM,}\\
\normalsize{$^{2}$University of Wuppertal, Chair of Reliability Engineering and Risk Analytics / IZMD}\\
\normalsize{{{\{}kowol, rottmann, bracke, hgottsch{\}}@uni-wuppertal.de.}}
}

\maketitle

\begin{abstract}
\textit{In this work, we present an uncertainty-based method for sensor fusion with camera and radar data.
The outputs of two neural networks, one processing camera and the other one radar data, are combined in an uncertainty aware manner. To this end, we gather the outputs and corresponding meta information for both networks. For each predicted object, the gathered information is post-processed by a gradient boosting method to produce a joint prediction of both networks.
In our experiments we combine the YOLOv3 object detection network with a customized $1D$ radar segmentation network and evaluate our method on the nuScenes dataset. In particular we focus on night scenes, where the capability of object detection networks based on camera data is potentially handicapped. Our experiments show, that this approach of uncertainty aware fusion, which is also of very modular nature, significantly gains performance compared to single sensor baselines and is in range of specifically tailored deep learning based fusion approaches.
}
\end{abstract}

\section{Introduction}
One of the biggest challenges for computer vision systems in automated driving is to recognize the cars environment appropriately in any given situation. The data of different sensors has to be interpreted correctly while operating with limited amount of computing resources. Besides dense traffic, different weather conditions like sun, heavy rain, fog and snow are challenging for state-of-the-art computer vision systems. Previous studies show that the use of more than one sensor, the so called sensor fusion, leads to an improvement in object detection accuracy, e.g.\ when combining camera and lidar sensors~\cite{DeSilva2018,Wu2017,Liu2017,Kragh2017, Gu2019}. Up to now, datasets containing real street scenes using radar data and another sensor are rather the exception, although synthetic data with different sensors, such as camera and radar sensors, can be generated using simulators like CARLA~\cite{dosovitskiy2017carla} or LSGVL~\cite{rong2020lgsvl}.
With the publication of the nuScenes dataset~\cite{Caesar2019}, the scientific community obtained access to real street scenes recorded with different sensors including radar. In total there are 5 radar sensors distributed around the car that generate the data. Ever since, the number of published papers dealing with sensor fusion combining radar with other sensors with the help of the nuScenes dataset has increased, see e.g.~\cite{John2019,Nobis2019,Perez2019}. We now briefly review these approaches. 

The authors of \cite{John2019} propose a convolutional neural network (CNN) for object detection named RVNet which is equipped with two input branches and two output branches. One input branch processes image data, the other one radar data. Similarly to YOLOv3~\cite{Redmon2018}, the network utilizes two output branches to provide bounding box predictions, i.e., one branch is supposed to detect smaller obstacles, the other one larger obstacles. The authors conclude that radar features are useful for detecting on-road obstacles in a binary classification framework. On the other hand the features extracted from radar data seem not to be useful in a multiclass classification framework due to the sparsity of the data.

Another deep-learning-based radar and camera sensor fusion for object detection is the CRF-Net (CameraRadarFusionNet)~\cite{Nobis2019}, which automatically learns at which level the fusion of both sensor data is most beneficial for object detection. The CRF-Net uses a so-called BlackIn training strategy and combines a RetinaNet (VGG backbone), a custom-designed radar network and a Feature Pyramid Network (FPN) for classification and regression problems. The main branch is composed of five VGG-Blocks, every block receives pre-processed radar and image data for further processing which is forwarded to the FPN-Blocks. The network is tested on the nuScenes dataset and a self-build one. The authors provide evidence that the BackIn training strategy leverages the detection score of a state-of-the-art object detection network.

Furthermore, a fusion approach for lidar and radar is introduced in \cite{Perez2019}. This approach is designed for multi-class object detection of pedestrian, cyclist, car and noise (empty region of interest) classes. To this end, lidar and radar data are first processed individually. The lidar branch detects objects and tracks them over time. On the other hand, the radar branch provides the object classification, where three independent fast Fourier transforms (FFTs) are applied on the range-Doppler-angle spectrum. After time synchronisation the two branches are merged, resulting in regions of interest. These regions are fed to the CNN, based on the VGGNet architecture, which computes the classes probabilities. 
Since this approach works well for vehicle and noise classification but has problems with pedestrians and cyclist classes the network was improved by applying a tracking filter on top of the classifier. They used a Bayes filter which improved the classification performance for the two challenging classes.

In summary, the works presented \cite{John2019,Nobis2019} aim at simultaneously fusing and interpreting image and radar data within a CNN. In \cite{Perez2019}, lidar and radar data are first fused and afterwards a CNN processes the fused input. While these approaches are beneficial with respect to maximizing performance, they require additional fallback solutions in case that a sensor drops out. 
Also in contrast to other sensor fusion solutions, our approach preserves the option to use both networks redundantly. In this way, indications of only one of the networks can be used for scenario constructions that are alternative to the main scenario provided by the fusion approach.

It also seems inevitable that sensor fusion approaches require additional uncertainty measures to verify the quality of the developed methods and networks. A tool for semantic segmentation called MetaSeg that estimates prediction quality on segment level was introduced in~\cite{Rottmann2018} and extended in  \cite{Maag2020, Rottmann19zoom, schubert2020metadetect}. It learns to predict whether predicted components intersects with the ground truth or not, which can be viewed as meta classifying between two classes ($\IoU = 0$ and $\IoU > 0$). To this end, metrics are derived from the CNN's output and pass them on to another meta-classifier. This work of false positive detection was extended in~\cite{Chan2020} where the number of overlooked objects was reduced by only paying with a few additional false positives. The overproduction of false positives is suppressed by MetaSeg. Following these approaches for uncertainty quantification, we use metrics from the output of two CNNs.
We pass them through to a gradient boosting classifier, which reduces the number of false positive predictions. In addition, by reducing the score threshold for object detection, we are able to improve over the performance of the respective single sensor networks.

In our tests, we utilize a YOLOv3 \cite{Redmon2018} as a state-of-the-art object detection network to process the camera data and complement this with a custom-designed CNN that performs a $1D$ binary segmentation which is supposed to detect obstacles. Further downstream of our computer vision pipeline we introduce a very general uncertainty-based fusion algorithm. Based on the predictions of both CNNs and their uncertainties as well as other geometrical meta-information, the fusion algorithm learns to provide a prediction by means of a structured dataset. In our experiments we demonstrate for the case of street scenes recorded at night, that this approach significantly improves the object detection accuracy. Furthermore, both networks only show moderate correlation which further supports our safety argument.

\medskip

\paragraph{\textbf{Outline}}
The remainder of this work is organized as follows. In \hyperlink{thesesentence}{\cref{sec:sensorfusion}} we briefly describe the characteristics of different sensors used for perception in automated driving. In \hyperlink{thesesentence}{\cref{detection_radar}} we introduce our $1D$ segmentation network for the detection of vehicles by radar data. We describe the preprocessing, the network architecture and the loss function we used for our method. This is followed by \hyperlink{thesesentence}{\cref{detection_via_fusion}}, where the sensor fusion approach is presented and in \hyperlink{thesesentence}{\cref{metrics}} our choice of metrics are introduced. In \hyperlink{thesesentence}{\cref{num_exp}} we discuss the numerical results. Finally, we present the conclusion and outlook in \hyperlink{thesesentence}{\cref{conclusion_outlook}}.

\begin{table}[ht!]
  \begin{center}
  \scalebox{0.7}{
    \begin{tabular}{|l|l|c|c|c|}
      \hline
      \multicolumn{2}{|c|}{\textbf{Specifications}} & \textbf{Camera}&\textbf{Radar}&\textbf{Lidar}\\
      \hline
      \multirow{2}{*}{\textbf{Distance}} & Range & ++ & +++ & +++ \\
      & Resolution & ++ & +++ & ++\\
      \hline
      \multirow{2}{*}{\textbf{Angle}} & Range & +++ & ++ & +++ \\
      & Resolution & +++ & + & ++\\
      \hline
      \multirow{2}{*}{\textbf{Classification}} & Velocity Resolution & + & +++ & ++ \\
      & Object Categorization & +++ & + & ++\\
      \hline
      \multirow{2}{*}{\textbf{Environment}} & Night Time & + & +++ & +++ \\
      & Rainy/Cloudy Weather & + & +++ & ++\\
      \hline
    \end{tabular}}
  \end{center}
  \centering + = Good, ++ = Better, +++ = Best\\
  \caption{Overview of the advantages and disadvantages of the most common sensors for autonomous driving\cite{Phillips2020}.}
  \label{tab:sensors}
\end{table}

\section{Sensor Characteristics} \label{sec:sensorfusion}
Today's vehicles are equipped with sensors for recording driving dynamics, which register movements of the vehicle in three axes, as well as sensors for detecting the environment. The latter try to map the vehicle environment as accurately as possible to promote automated driving. This section briefly describes the characteristics of the three most used sensors for the perception of the environment for automated driving, i.e., camera, radar and lidar, including their advantages and disadvantages.

\begin{figure*}[ht]
    \centering
    \includegraphics[width=\textwidth, trim={0 4cm 0 5cm},clip]{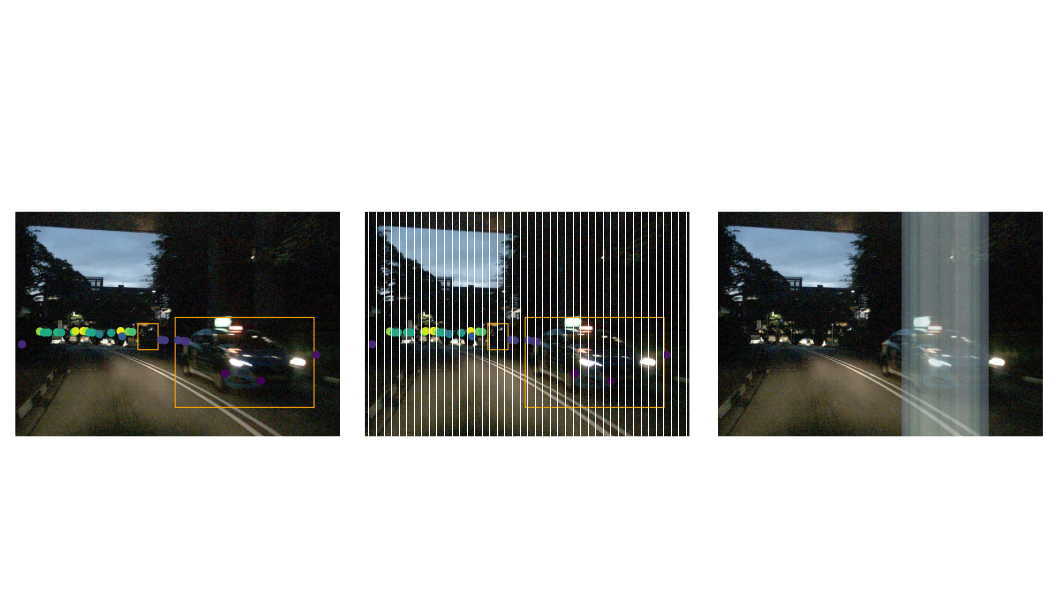}
    \caption{Preprocessing and prediction of the radar network. On the left we see the radar points projected to the front view camera image. The center image is divided into a certain number of slices $N_s$ from which we generate the input matrix for the training of the CNN. The right hand image corresponds to the output of the radar network.}
    \label{fig:process_cnn_radar}
\end{figure*}

Camera sensors take two-dimensional images of light by electrical means. They are accurate in measuring edges, contour, texture and coloring. Furthermore they are easily integrated into the design of a modern vehicle. However, $3D$ localization from images is challenging and weather-related visual impairment can lead to higher uncertainties in object detections \cite{Wang2020}.

Radar sensors use radio waves to determine the range, angle and relative velocity of objects. Long-range radar sensors have a high range capability up to $200-250m$ \cite{John2019,Schneider2005,Mueller2017} and are cheaper than Lidar sensors~\cite{Aldrich2018}. Compared to cameras, radar sensors are less affected by environmental conditions and pollution~\cite{Aldrich2018,Fritsche2016, Mueller2017}. On the other hand, radar data is sparse and does not delineate the shape of the obstacles \cite{Fritsche2016, John2019}. 

Lidar sensors use a light beam, emitted from a laser, to determine the distances and shapes of objects.
Lidar sensors are highly accurate in $3D$ localization and surface measurements as well as a long-range view up to $300m$~\cite{Pidurkar2019}. They are expensive to buy and bad weather conditions like rain, fog or dust reduce the performance \cite{Aldrich2018, Mueller2017}.

Each sensor has its advantages and disadvantages (summarized in \cref{tab:sensors}), so that a sensor fusion with at least two sensors makes sense in order to provide a better safety standard.

\section{Object Detection via Radar} \label{detection_radar}

In this section we introduce the $1D$ segmentation network that we equip for detecting vehicles. First we explain the pre-processing method, then we describe the network architecture, the loss function and the network output.

\subsection{Preprocessing} 
In a global $3D$ coordinate system, the radar data is situated in a $2D$ horizontal plane. Hence, before training a neural network, we pre-process the radar data for two reasons. First of all, in order to simplify the fusion after processing each sensor with a neural network separately, we project the given radar data into the same $2D$ perspective as given by the front view camera. Secondly, one can observe that after this projection, the remaining section of the radar sensor modality is close to $1D$.
\Cref{fig:process_cnn_radar} depicts radar points projected  into the front view camera image. Darker colors indicate closer objects and brighter colors indicate more distant objects.
Consequently, we build and train a neural network to perform a $1D$ segmentation.

\begin{figure}[ht]
    \centering
    \includegraphics[width=.45\textwidth, trim={0 1.5cm 0 0}, clip]{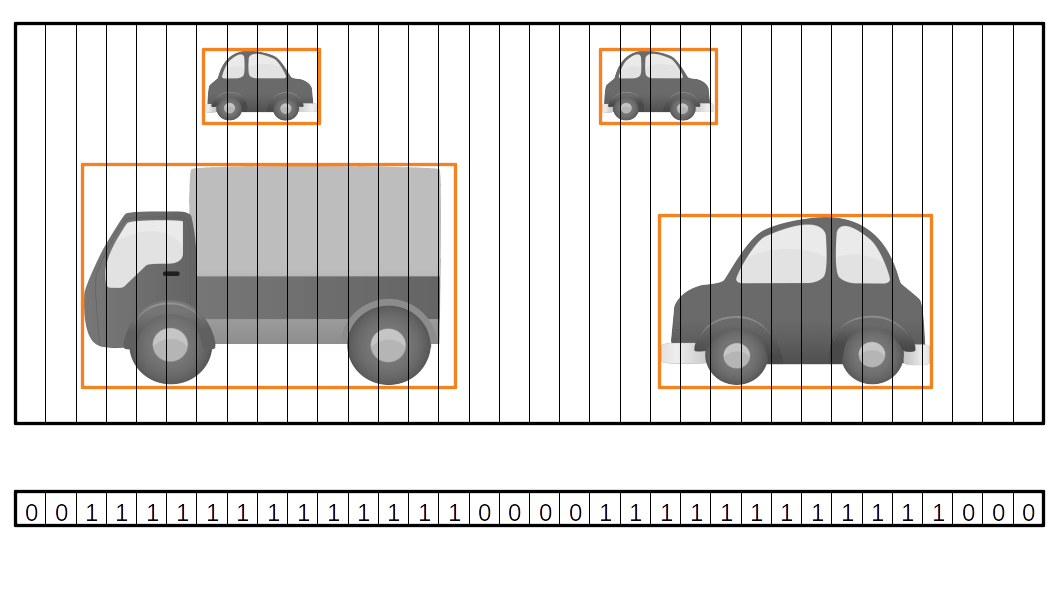}
    \caption{Ground truth vector for radar data. Every slice that overlaps with a bounding box in the front view camera image obtains the value 1, otherwise 0.}
    \label{fig:gt_vector}
\end{figure}

To be more specific, we pre-process the ground truth for training the radar network as illustrated in \cref{fig:gt_vector}. That is, we divide the given front view image into a chosen number $N_s$ of slices and generate an occupancy array of length $N_s$. The $i$th entry of this array is equal to $1$ if there is a ground truth object intersecting with the $i$th slice of the image and $0$ else.
This ground truth construction defines the desired prediction for the radar network.
\begin{figure*}[ht]
    \centering
    \includegraphics[width=\textwidth, trim={0 2.5cm 0 0}, clip]{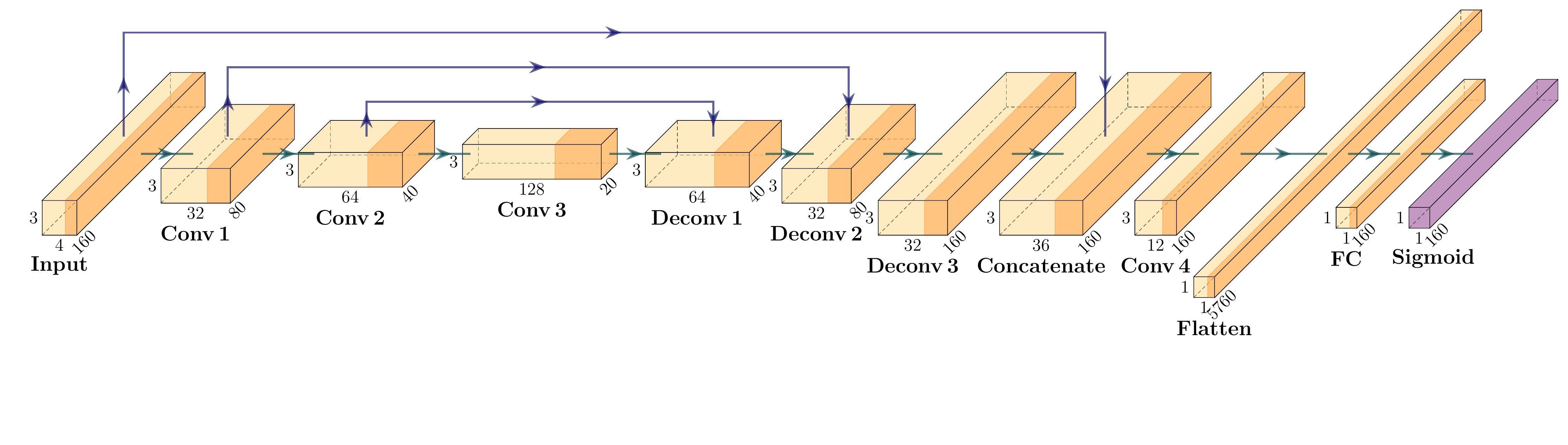}
    \caption{CNN architecture of our custom FCN-8 inspired radar network.}
    \label{fig:architecture}
\end{figure*}

The radar data is pre-processed similarly to the ground truth as we aim at providing the neural network with an input tensor of size $N_s \times N_t \times N_f$ where $N_t$ denotes the number of considered time steps and $N_f$ denotes the number of features. By assigning radar points to image slices we can drop the $x$-coordinate (which is implied by the array index up to an quantization error). For each slice $i=1,\ldots,N_s$ which contains at least one radar point we store the following $N_f$ features in the input matrix: $y$-coordinates (indicating the distance of the reflection point), the height coordinate with respect to the front view image that the radar point obtains by projection into the image plane as well as the relative lateral and the longitudinal velocity.

\subsection{Network Architecture}
The network architecture for the radar network is based on a FCN-8-network~\cite{Long2015} and is depicted in \cref{fig:architecture}. The hidden layers consist of three convolution blocks, three deconvolution blocks followed by a concatenate layer, a fourth convolution block, a flatten layer and one fully-connected block. Finally, the network contains a sigmoid layer from which we get in $[0,1]$. Each convolution and deconvolution block includes a (de-)convolution layer, a batch normalization and a leaky ReLU as activation function, respectively. The convolution blocks capture context information while losing spatial information whereas the deconvolution blocks restore these spatial information. Using bypasses, context information can be linked with spatial information. Furthermore, each fully connected block consists of a dense layer followed by a leaky ReLU activation function (except for the final layer where we use a sigmoid activation function).

\subsection{Loss function}

Let $D = \{ (r^{(i)},t^{(i)}) : i = 1,\ldots,n \}$ denote a dataset of tuples containing radar data $r^{(i)} \in \mathbb{R}^{N_s \times N_t \times N_f}$ and ground truth $t^{(i)} \in \{0,1\}^{N_s}$. The radar network $g$ provides an array of estimated probabilities indicating whether a given slice $s$ is occupied or not. We denote $y^{(i)} = g(r^{(i)})$.
For training the neural network we use the binary cross-entropy for each array entry  $s=1,\ldots,N_s$, i.e., for a single data sample $(r^{(i)},t^{(i)})$ we have 
\begin{equation}
    \ell(t^{(i)}_s,y^{(i)}_s) = - \alpha t^{(i)}_s \log( y^{(i)}_s ) - ( 1- t^{(i)}_s) \log( 1- y^{(i)}_s )
\end{equation}

where $\alpha$ is a tunable parameter. We introduced this parameter in order to account for the imbalance of zeros and ones in the ground truth. When training with stochastic batch gradient descent, the loss function is summed over all  $s=1,\ldots,N_s$ and then the mean is computed over all indices $i$ in the batch.

\begin{figure}[ht!]
    \centering
    \includegraphics[width=.54\textwidth, trim={8cm 3cm 3cm 4cm}, clip]{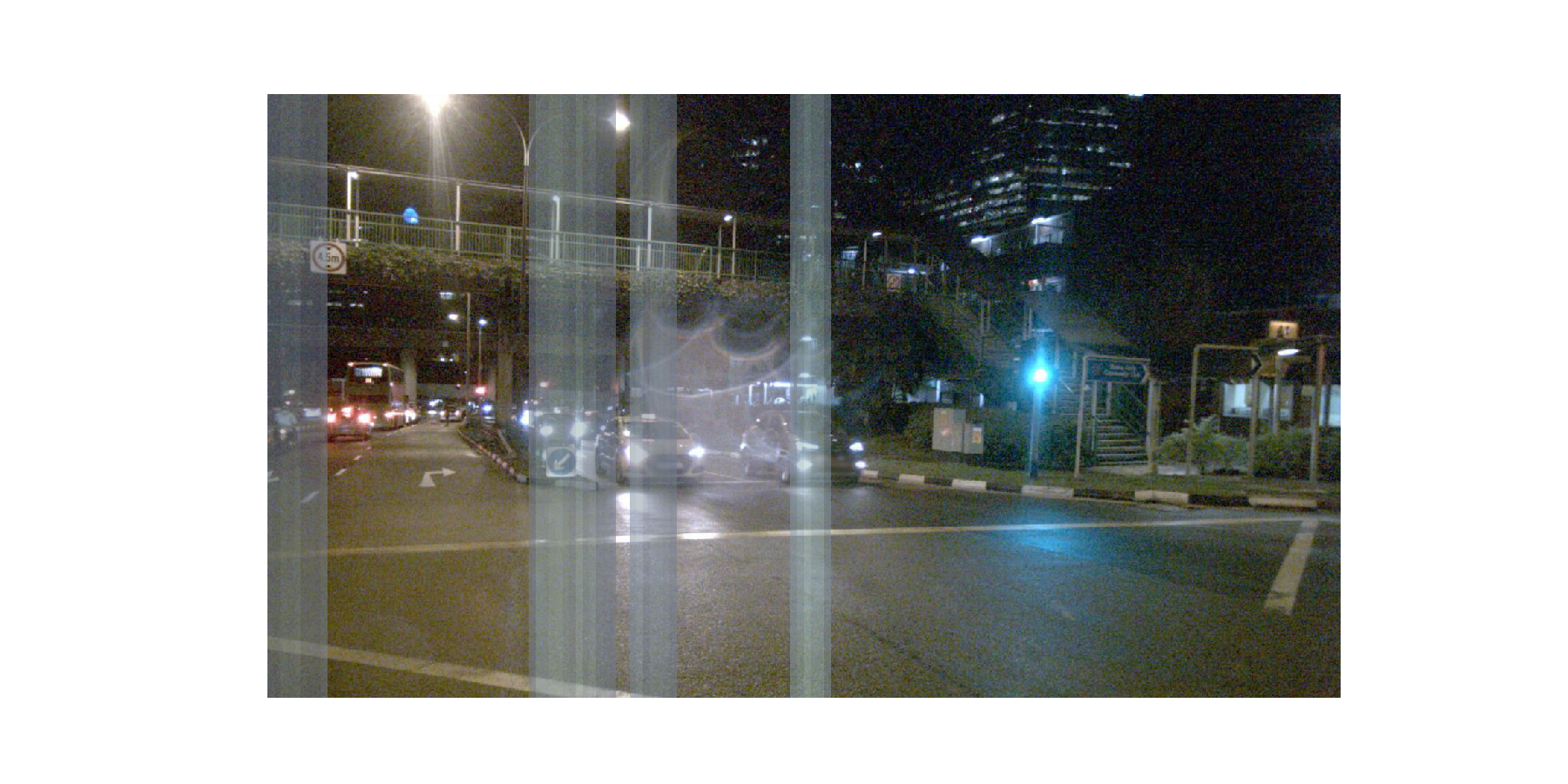}
    \caption{Image of a radar detection example with four predicted slice bundles}
    \label{fig:ex_radar}
\end{figure}

\begin{figure*}[ht]
    \includegraphics[width=\textwidth]{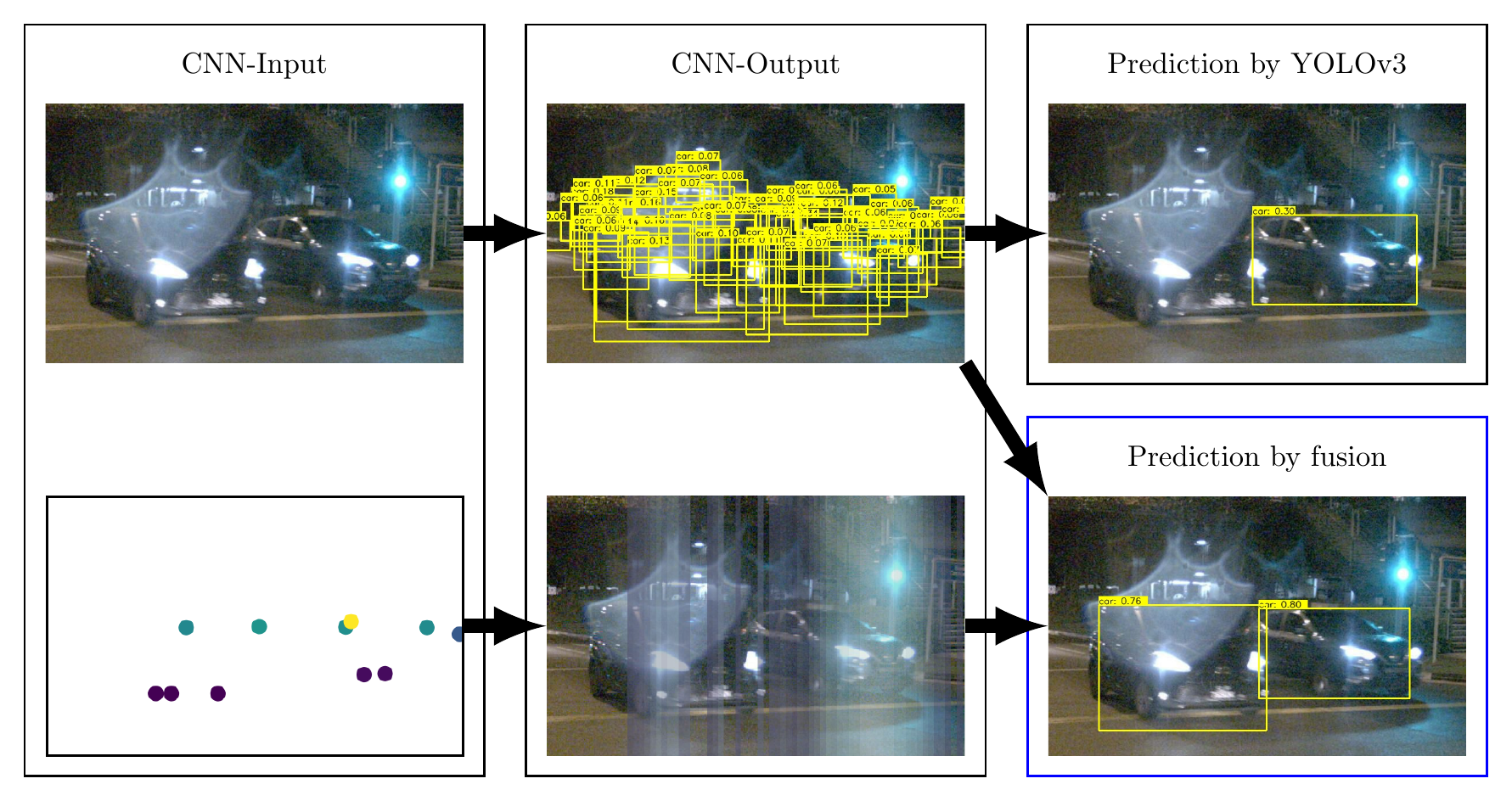}
    \caption{Illustration of our YOdar method. In the top branch, the YOLOv3 produces many candidate bounding boxes, but after score thresholding and non maximum suppression, only one of both cars is detected. By lowering the object detection threshold and fusing the obtained boxes with the radar prediction, both cars are detected by YOdar as shown in the bottom right panel.}
    \label{fig:yodar}
\end{figure*}

\subsection{Output}
The predictions of the radar network result in a vector consisting of values in $[0,1]$ that estimate the probability of occupancy. Neighboring slices whose predicted probabilities are above a certain threshold $T_g$ are recognized as one coherent object, also called slice bundle. \Cref{fig:ex_radar} shows for example an image with four slice bundles. The more a slice bundle fills in a bounding box, the higher the $1D$ $\IoU$ gets. 

\section{Object Detection via Sensor Fusion}\label{detection_via_fusion}
After describing the object detection method using radar sensor data in the previous section, this section deals with the image detection method and the fusion of both methods. Various object detection networks have been developed in recent years, whereby the YOLOv3 network has become a very good choice when fast and accurate real-time detection is desired~\cite{Redmon2018, Benjdira2019}. It has been observed that YOLOv3 works very well under good weather and visibility conditions but has problems with object recognition in bad visibility like hazy weather~\cite{Tumas2020, Li2020} or darkness, like Xiao et al. have investigated for object detection with RFB-Net~\cite{Xiao2020}. In our experiments, we focus on object detection by camera and radar at night. To this end, we first use each method separately in order to connect both outputs with gradient boosting~\cite{Friedman2002}, see \cref{fig:yodar}. 
Similarly to~\cite{Rottmann2018} we derive metrics from each CNN output and pass them through a gradient boosting classifier to increase the number of detected vehicles. The data and metrics used for the classifier are explained in \cref{metrics}.
The YOLOv3 threshold $T_f$ for vehicle detection is set to a low value such that we get a higher number of bounding box predictions. From the many predicted bounding boxes, gradient boosting select those boxes that are likely to contain an object according to the output of both networks. On the one hand, the radar sensor should detect vehicles not recognized by the YOLOv3 network. On the other hand, gradient boosting should support the decision making process by additional information in case the YOLOv3 network is uncertain.

\section{Fusion Metrics and Methods} \label{metrics}

The fusion method that we introduce in this section is of generic nature.
Therefore, we denote by $f$ the an arbitrary camera network and by $g$ a $1D$ segmentation radar network.
Given an input scene $(x,r)$, we obtain two network outputs, one for the image input $x$, one for the radar input $r$. Each prediction obtained by $f$ consists of a set $B = \{ b_1, \ldots, b_{k} \}$ containing $k$ boxes where $k$ depends on $x$.
Each box $b_i$ is identified with a tuple $\beta_i$ that contains an objectness score value $z_i$, a probability $f(v|x,b_i)$ that the box $b_i$ contains a vehicle $v$, a center point with its x-coordinate $c_i^x \in \mathbb{R}$ and y-coordinate $c_i^y \in \mathbb{R}$ as well as width $w_i \in \mathbb{R}$ and height $h_i \in \mathbb{R}$ of the box, i.e.,

\begin{equation}
    \beta_i = ( z_i, f(v|x,b_i), c_i^x, c_i^y, w_i, h_i ) \, .
\end{equation}

For the radar network $g$ we obtain a $1D$ output of probabilities, $g_s( v | r )$ for each of the slices $s = 1,\ldots,n $, that this slice $s$ belongs to a vehicle $v$, recall \cref{fig:gt_vector}.
As depicted in \cref{fig:process_cnn_radar} we identify slices $s$ and bounding boxes $b_i$.
In order to aggregate slices $s$ from the radar network over bounding boxes $b_i$ obtained by the camera network, let $S_i$ denote the set of all slices $s$ that intersect with the box $b_i$. We denote by
\begin{equation} \label{eq:radarmeanoverbox}
    \mu_i  = \frac{1}{| S_i |} \sum_{s \in S_i} g_s( v | r ) \, 
\end{equation}
the average probability of observing a vehicle in the box $b_i$ according to the radar network's probabilities. The standard deviation corresponding to \cref{eq:radarmeanoverbox} is termed $\sigma_i$. As a set of metrics, by which we compute a fused prediction, we consider
\begin{equation}
M_i(x,r) = ( \beta_i, A_i, \mu_i, \sigma_i )
\end{equation}
where $A_i = w_i \cdot h_i$ denotes the size of the box $b_i$. In summary, we use these nine metrics $M_i(x,r)$ for all scenes $(x,r)$ and boxes $b_i$ that are visible with respect to the front view camera.
To perform the fusion of the camera based network prediction and the radar based network prediction we proceed in two steps. First we compute the ground truth which states for each box predicted by the camera network whether it is a true positive (TP) or a false positive (FP). Afterwards we train a model to discriminate by means of $M_i$ whether $b_i$ is a TP or an FP.

More precisely, for the sake of computing ground truth, we define TP and FP in the given context as follows: For a predicted box $b_i$ and a ground truth box $a$, which has the biggest intersection $|a \cap b_i|$ of all ground truth boxes of the same class, the intersection over union is defined as follows:
\begin{equation}
    \IoU(b_i) = \max_{a} \frac{|a \cap b_i|}{|a \cup b_i|} \, .
\end{equation}
Oftentimes we omit the argument $b_i$ if it is clear from the context. Given a chosen threshold $T \in [0,1)$ we define that $b_i$ is a TP if $\IoU(b_i) > T$ and an FP if $\IoU(b_i) \leq T$. For the sake of completeness, we define that a false negative is a ground truth box $a$ that fulfills $\max_{b_i} \frac{|a \cap b_i|}{|a \cup b_i|} \leq T $. Note that in the latter definition, the ground truth element is fixed while the left hand side of this expression is maximized over all predicted boxes.

After computing the ground truths, i.e., whether a box $b_i$ predicted by the camera network yields a TP or an FP, we train a model. The gathered metrics $M_i$ yield a structured dataset where the columns are given by the different metrics and the rows are given by all predicted boxes $b_i$ for each input scene $(x,r)$. By means of this dataset and the corresponding TP/FP annotation, we train a classifier to predict whether a box predicted by the camera network is a TP or an FP. 

\begin{figure}[h!]
    \centering
    \includegraphics[width=.4\textwidth]{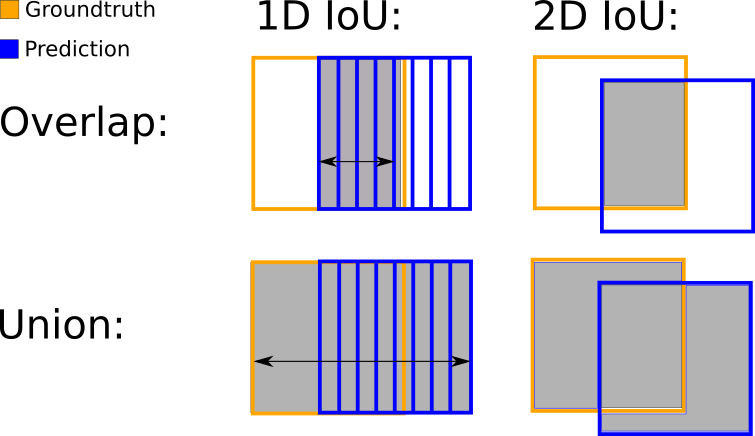}
    \caption{$\IoU$ calculation for $1D$ and $2D$ bounding boxes}
    \label{fig:iou}
\end{figure}

The $\IoU$ can be calculated in different dimensions $D=1,2$. In this work the $1D$ $\IoU$ is used for the radar network. As soon as a low threshold $T_g$ has been reached, a predicted object is considered as TP, otherwise as FP. The $2D$ $\IoU$ is used for the YOLOv3 network, analogously we speak of TP and FP according to a threshold $T_f$. An illustration is given in \cref{fig:iou}.

The mean average precision ($\mAP$) is a popular metric used to measure the performance of models. The $\mAP$ is calculated by taking the average precision (area under precision as a function of recall, a.k.a.\ precision recall curve) over one class.

\section{Numerical Experiments} \label{num_exp}
As explained in detail in the previous section, we use a custom FCN-8-like network for processing the radar data from the nuScenes dataset~\cite{Caesar2019}. It contains urban driving situations in Boston and Singapore. The dataset has a high variability of scenes, i.e., different locations, weather conditions, daytime, recorded with left- or right-hand traffic. In total, the dataset contains $1,\!000$ scenes, each of $20$ seconds duration and each frame is fully annotated with $3D$ bounding boxes. The vehicle used, a Renault Zoe, was set up with $6$ cameras at $12Hz$ capture frequency, $5$ long-range radar sensors (FMCW) with $13Hz$ capture frequency, $1$ spinning lidar with $20Hz$ capture frequency, $1$ global positioning system module (GPS) and $1$ inertial measurement unit (IMU). Each scene is divided into several time frames for which each sensor provides a suitable signal. 

\begin{table}[ht!]
  \begin{center}
  \scalebox{0.7}{
    \begin{tabular}{|c||C{1.9cm}|C{1.9cm}||C{1.9cm}|C{1.9cm}|}
      \hline
      \multirow{2}{*}{\textbf{splitting}} & \multicolumn{2}{c||}{\textbf{number of images/frames}} & \multicolumn{2}{c|}{\textbf{night images/frames [\%]}} \\
       & \textbf{YOLOv3}&\textbf{Radar} & \textbf{YOLOv3}&\textbf{Radar}\\
      \hline
      train & $10,\!000$ &  $29,\!853$ & $7.08$ &  $9.04$ \\
      \hline
      val &  $3,\!289$ & $3,\!289$ &  $8.57$ &  $8.57$ \\
      \hline
      test & $1,\!006$ & $1,\!006$ & $100$ & $100$ \\
      \hline
    \end{tabular}}
  \end{center}
  \caption{Data used for training, validation and testing. }
  \label{tab:used_data}
\end{table}

For our experiments, the YOLOv3 network was pretrained with day images from the COCO dataset \cite{Lin2014} and afterwards with $249$ randomly selected scenes from the nuScenes dataset containing $10,\!000$ images with different weather conditions and times of day. Furthermore we have trained a CNN with radar data from the nuScenes dataset to detect vehicles. $743$ of the scenes ($29,\!853$ frames) were used as training data, $82$ scenes ($3,\!289$ frames) as validation data and $25$ scenes ($1,\!006$ frames) as test data, see \cref{tab:used_data}. For the test set, we consider exclusively all frames recorded at night that are not part of the training data in order to create a perception-wise challenging test situation. For the training and validation sets we used a natural split of day and night scenes pre-defined by the frequencies in the nuScenes dataset. Due to the resolution of the radar data, we focus on the category vehicle in our evaluation. This includes the semantic categories car, bus, truck, bicycle, motorcycle and construction vehicle.

\subsection{Training}

As input, the radar network obtains a tensor with the dimensions $160 \times 3 \times 4$ that contains for each of the $160$ considered slices the current frame (i.e., the current time step) and two previous frames. Each frame contains four features, i.e., x-, and y-coordinates, lateral and longitudinal velocity.
From the radar data we removed all ground truth bounding boxes that do not contain any radar points with valid velocity vectors. 

\begin{figure*}[ht]
\centering
    \includegraphics[width=\textwidth]{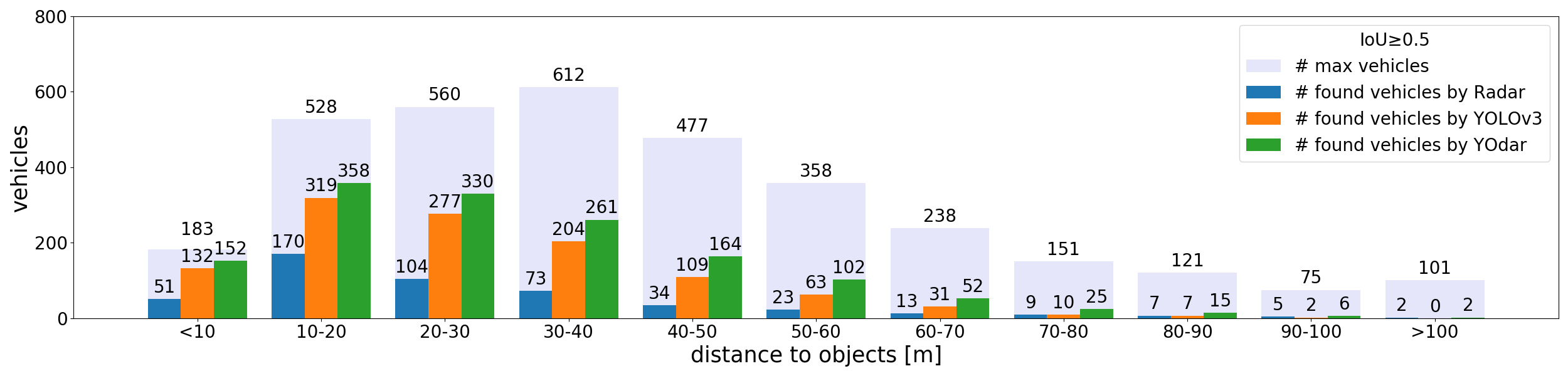}
    \caption{Vehicles recognized at night with consideration of the respective distance from the ego car. An object is considered as detected when it has an $\IoU \geq 0.5$ with the ground truth. These are average values from three test runs.}
    \label{fig:distances}
\end{figure*}

The radar network is implemented in Keras~\cite{Chollet2015} with TensorFlow~\cite{tensorflow2015-whitepaper} backend. Training on one NVIDIA Quattro GPU P6000 takes $229$ seconds training time. The network structure is shown in \cref{fig:architecture} and the training parameters are shown in \cref{tab:para}. We have trained the neural network three times with three different learning rates\, i.e., the first $20$ epochs with a learning rate of $10^{-3}$, $10$ epochs with $10^{-4}$ and 10 epochs with $10^{-5}$. The networks output vector has the same dimension $(160 \times 1)$ as the ground truth vector, where each entry contains a probability value, whether there is a vehicle in the respective area or not. If the probability value of a single slice is equal or higher than the threshold $T_g=0.5$, then the network predicts a vehicle. The higher the probability value, the brighter the slice is displayed in \cref{fig:process_cnn_radar}.

\begin{table}[h!]
  \begin{center}
  \scalebox{0.7}{
    \begin{tabular}{|c||C{1.3cm}|C{1.3cm}|C{1.3cm}|c|}
      \hline
      \textbf{Parameter}& \multicolumn{3}{c|}{ \textbf{Radar}} &\textbf{YOLOv3}\\
      \hline
      Batchsize & \multicolumn{3}{c|}{128} & 6  \\
      \hline
      Learning rate & \centering $10^{-3}$ & \centering $10^{-4}$ & \centering $10^{-5}$ & $10^{-4}$ - $10^{-6}$ \\
      \hline
      Epochs & \centering 20 & \centering 10 & \centering 10 & 100 \\
      \hline
      Weight decay &\multicolumn{3}{c|}{$3\cdot10^{-4}$} & variable \\
      \hline
      Loss function & \multicolumn{3}{c|}{modified binary-crossentropy} & binary-crossentropy\\
      \hline
      Optimizer & \multicolumn{3}{c|}{Adam} & Adam \\
      \hline
    \end{tabular}}
  \end{center}
  \caption{Training Parameters.}
  \label{tab:para}
\end{table}

The YOLOv3 network was trained with $10,\!000$ images consisting of different scenes. 
Therefore we converted the $3D$ bounding boxes in the nuScenes dataset into $2D$ bounding ones. To this end, we chose the smallest $2D$ bounding box that contains the front and rear surfaces of the $3D$ bounding box in the given ego car view. YOLOv3 is implemented in the Python-Framework Tensor-Flow~\cite{tensorflow2015-whitepaper}. Training with the same GPU as used for the custom radar network takes $50,\!32$ hours training time. The training parameters are stated in~\cref{tab:para}.

\subsection{Evaluation}
All results in this section are averaged over $3$ experiments to obtain a better statistical validity. \Cref{fig:distances} shows absolute numbers of recognized objects ($\IoU \geq 0.5$ with the ground truth) at night for each of the networks (Radar and YOLOv3) standalone as well as for our uncertainty-based fusion approach (YOdar). The numbers are broken down corresponding to distance intervals along the horizontal axis.
The lavender bar (in the background) displays the numbers of vehicles in the ground truth for the given distance interval. The blue bar states the numbers of objects recognized by the radar network. The performance of the radar network is low, only a small percentage of objects are found. After 20 meters, the performance decreases with growing distance.
The poor recognition of objects from radar can be explained by the small number of points provided for each frame. In addition, relative velocities are used for training, which means that mainly moving objects can be recognized and stationary or parked vehicles remain undetected.
The orange bar shows a significant increase of objects recognized by the YOLOv3 network compared to the radar network. Although mainly closer objects are recognized, there remain difficulties in object recognition with more distant objects. The green bar shows the objects recognized by YOdar. Compared to the YOLOv3 network, more vehicles are recognized for each of the given distances. In total, compared to the YOLOv3 network, the sensor fusion approach recognizes $313$ vehicles more (which amounts to an increase of $9.20$ percent points).

So far, we have seen that we recognize more objects with the YOdar approach than with YOLOv3 or our radar network separately. However, the increased sensitivity also yields some additional FPs. In order to compare the number of FPs for YOLOv3 and YOdar, we adjust the sensitivity of the YOLOv3 network by lowering the threshold $T_f$ such that the TP level for YOLOv3 is roughly equal to the TP level of YOdar. The resulting number of FPs is given in~\cref{tab:fp_study}. Indeed, YOdar generates $575$ less FP predictions than YOLOv3 for a common TP level, on which we let YOdar operate in our tests.

\begin{table}[ht]

    \begin{center}
    \scalebox{0.7}{
    \begin{tabular}{|c||C{1.5cm}|C{1.5cm}||C{1.5cm}|C{1.5cm}|}
        \hline
        \textbf{Network} & \multicolumn{2}{c||}{\textbf{unchanged output}} & \multicolumn{2}{c|}{\textbf{TP level adjustment}}\\
         & TP & FP & TP & FP \\
        \hline 
        YOLOv3 & $1,\!154$ & $98$ & $1,\!478$ & $1,\!024$ \\
        YOdar & $1,\!467$ & $449$ & $1,\!467$ & $449$ \\
        \hline 
    \end{tabular}}
    \end{center}
    \caption{Comparison of the number of false positives for YOLOv3 and YOdar at a common level of false positives.}
    \label{tab:fp_study}
\end{table}

\begin{figure}[ht!]
\centering
    \includegraphics[width=.5\textwidth, trim={9cm 0 2cm 0},clip]{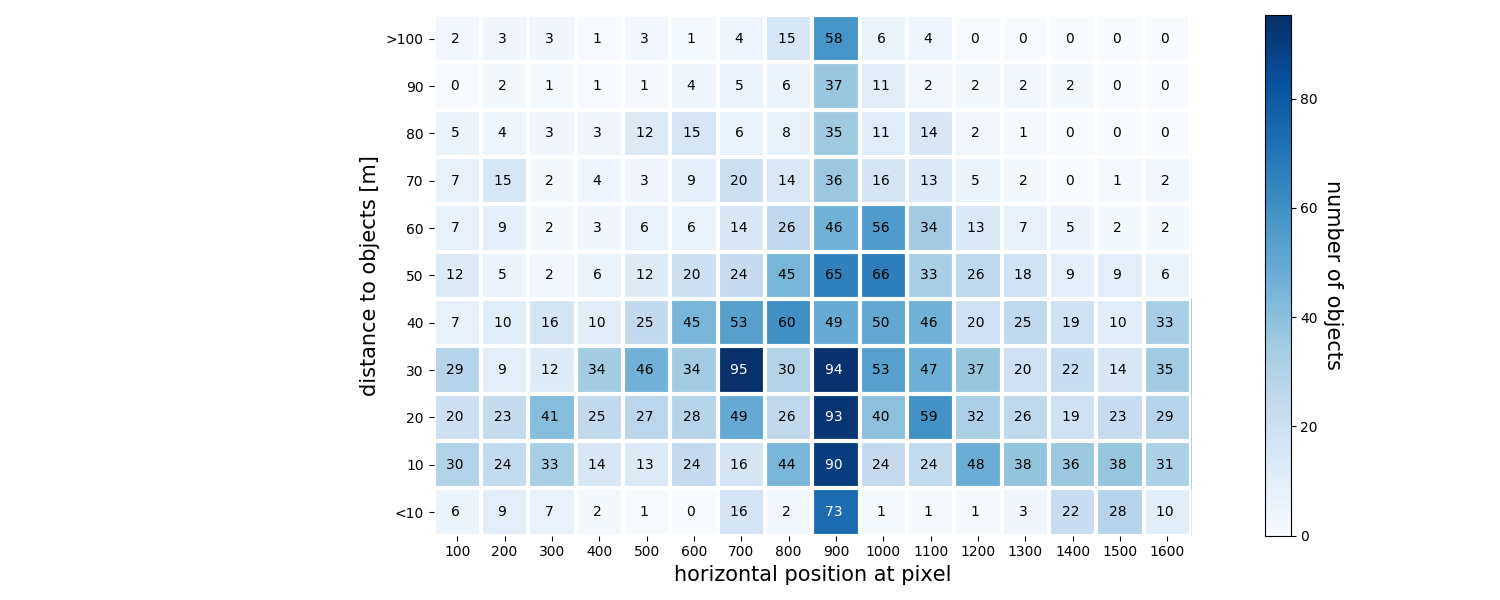}
    \caption{Ground truth heat map displaying the spatial distribution of the test data, broken down by distance (vertical axis) and the pixel intervals of width 100 of the front view input image with a total width of 1600 pixels (horizontal axis). More precisely, each intervals denoted by $i$ on the horizontal axis represents the pixels $(\mathit{row},\mathit{column})$ with $\mathit{column} \in [i-99,i]$.}
    \label{fig:heatmap_gt}
\end{figure}

Digging deeper into the discussed results, we now break down the distance intervals along the distance radii.
\Cref{fig:heatmap_gt} states the absolute numbers of objects broken down by distance (vertical axis) and the pixel intervals of width 100 of the input image with a total width of 1600 pixels (horizontal axis). More precisely, each interval denoted by $i$ on the horizontal axis represents the pixels $(\mathit{row},\mathit{column})$ with $\mathit{column} \in [i-99,i]$. A ground truth object is a member of such an interval, if the center of the box is contained in the respective interval and has the respective distance from the ego car. Thus, this can be viewed as a spatial distribution of the ground truth where the center of the bottom row is the area closest to the ego car.
The majority of the objects is located in the intervals given by $i=500,\ldots,1,\!200$.

\begin{figure}[ht!]
\centering
    \includegraphics[width=.5\textwidth, trim={9cm 0 2cm 0},clip]{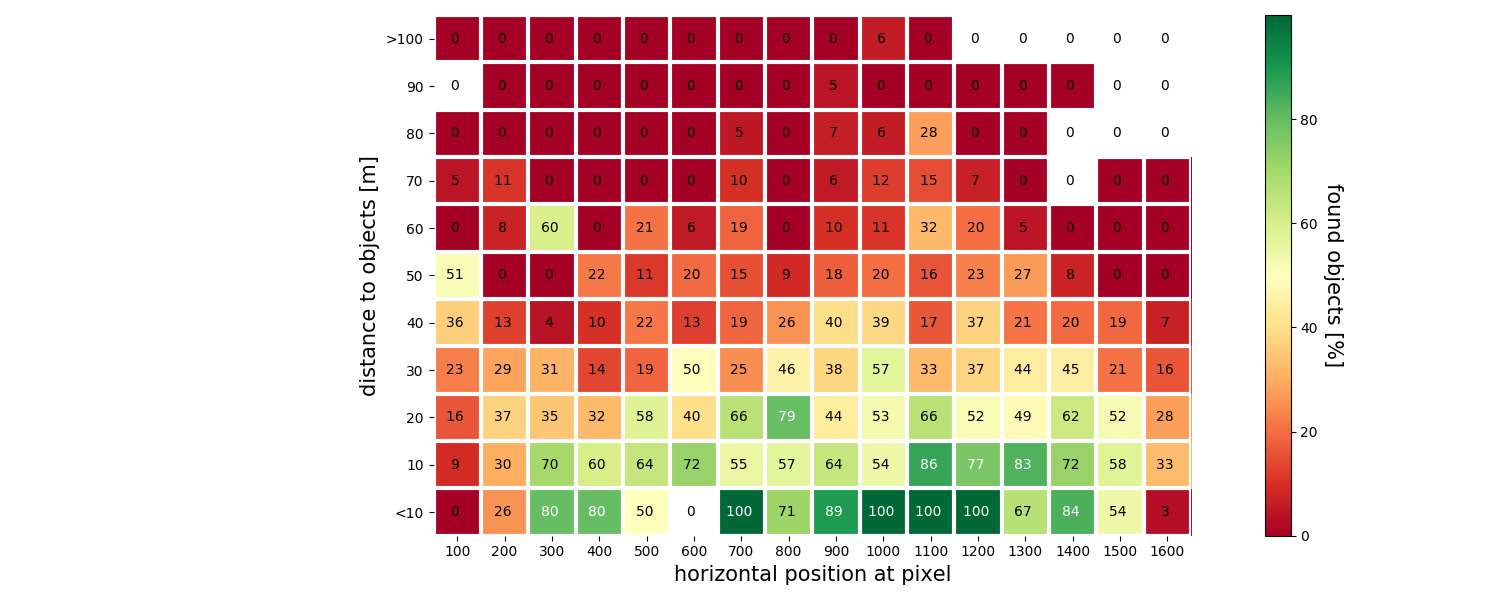}
    \caption{Relative amount of objects recognized by the YOLOv3 network evaluated on the test data. The underlying geometry is the same as in \cref{fig:heatmap_gt}.}
    \label{fig:heatmap_yolo}
\end{figure}

\begin{figure}[ht!]
\centering
    \includegraphics[width=.5\textwidth, trim={9cm 0 2cm 0},clip]{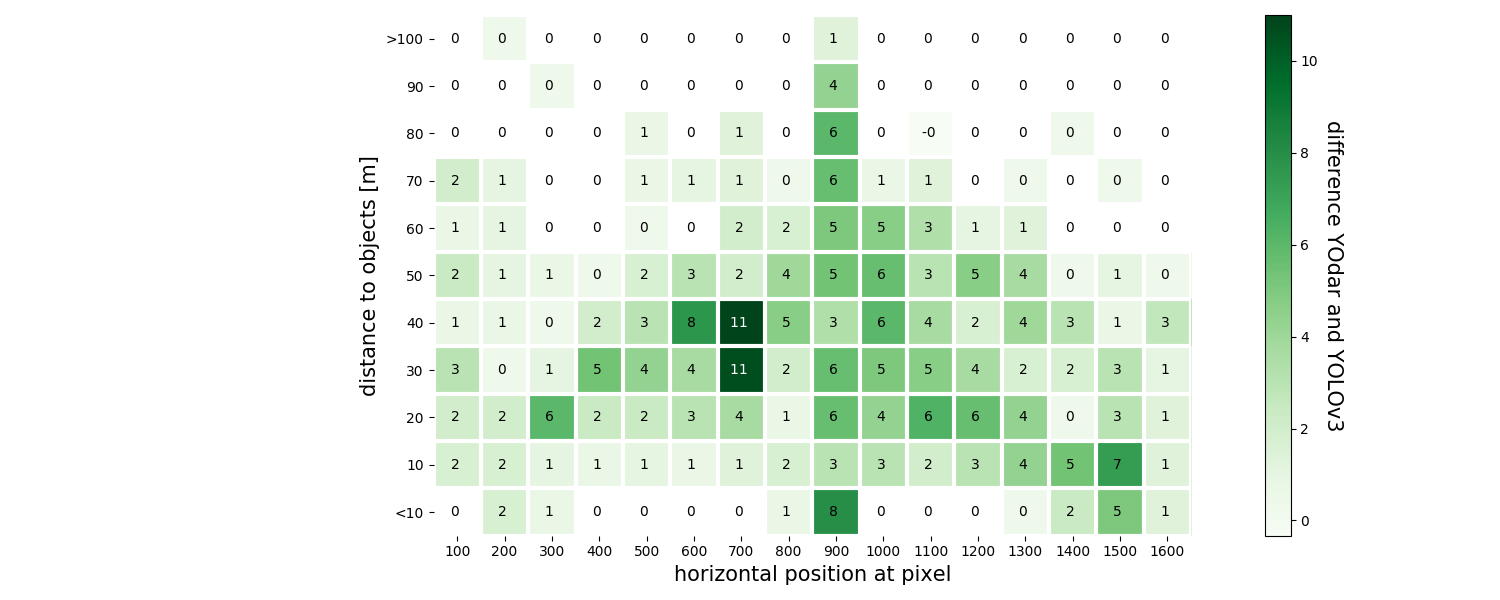}
    \caption{Number of objects recognized by YOdar minus the number of objects recognized YOLOv3. The underlying geometry is the same as in \cref{fig:heatmap_gt}.}
    \label{fig:heatmap_change}
\end{figure}

\Cref{fig:heatmap_yolo} shows the relative amount of objects recognized by the YOLOv3 network. It shows that mainly objects closer to the ego car and straight ahead are recognized, while objects farther away or located on the very left or very right end of the image remain often unrecognized. 
\Cref{fig:heatmap_change} states in absolute numbers how many additional objects are recognized in each particular area when using YOdar instead of YOLO. The increase is clear and also mostly in the relevant areas close to the ego car and straight ahead. This is in line with the idea of focusing with the radar on objects in motion (by considering objects that carry velocities). These results show that an uncertainty-based fusion approach like YOdar is indeed able to increase the performance significantly. This finding is also confirmed by the $\mAP$ and accuracy values stated in \cref{tab:results}. While YOLOv3 achieves $31.36\%$ $\mAP$, Yodar achieves 39.40\% which is also close to state of the art deep learning based fusion results for the nuScenes dataset with the natural split of day and night scenes as reported in~\cite{Nobis2019}.

\begin{table}[h!]
  \begin{center}
  \scalebox{0.7}{
    \begin{tabular}{|c||c|c|c|}
      \hline
      \textbf{}& \textbf{Radar}&\textbf{YOLOv3}&\textbf{YOdar}\\
      \hline
      Accuracy [\%] & 14.42 & 33.90 & 43.10  \\
      \hline
      mAP [\%] & 7.93  & 31.36 & 39.40 \\
      \hline
    \end{tabular}}
  \end{center}
  \caption{Accuracies and $\mAP$ scores of all three networks.}
  \label{tab:results}
\end{table}

\begin{figure*}[ht!]
    \centering
    \includegraphics[width=\textwidth]{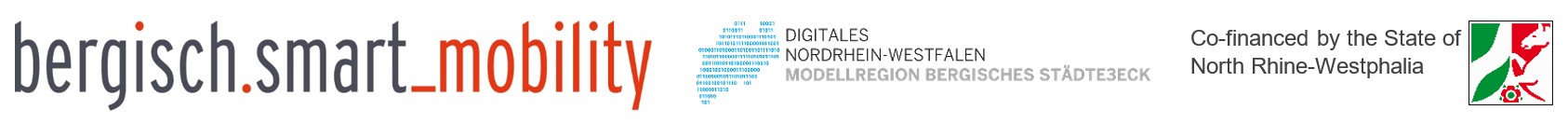}
    \label{fig:logo_bsm}
\end{figure*} 

\section{Conclusion and Outlook} \label{conclusion_outlook}
In this paper we have introduced the method YOdar which detects vehicles with camera and radar sensors. With this uncertainty-based sensor fusion approach the camera and radar data are first processed individually. Each branch detects objects, the camera branch uses a YOLOv3 network trained with day and night scenes and the radar branch uses a custom-based radar network. The outputs of every branch are aggregated and then passed through a post processing classifier that again learns the same vehicle detection task. Compared to the YOLOv3 network, the YOdar fusion method detects at night a significant additional amount of vehicles in total. While YOLOv3 achieves $31.36\%$ $\mAP$, YOdar achieves $39.40\%$ $\mAP$ which is also close to state of the art deep learning based fusion results for the nuScenes dataset with the natural split of day and night scenes.
In future work, additional sensors will be added for training and evaluation, since this approach uses only the front camera and the front radar sensor. Furthermore, we plan to optimize the radar network such that more objects can be detected by YOdar. Moreover, we plan to extend this approach also the detection of pedestrians as more dense radar data becomes available.

\section*{Acknowledgements}
K.K.\ acknowledges financial support through the research consortium bergisch.smart.mobility funded by the ministry for economy, innovation, digitalization and energy (MWIDE) of the state North Rhine Westphalia under the grant-no.\ DMR-1-2. 

\bibliography{citation}
\bibliographystyle{unsrt}

\end{document}